\documentclass[acmtog]{acmart}
\AtBeginDocument{%
  }

\usepackage{multirow}
\usepackage{graphicx}
\usepackage{pifont}
\usepackage{enumitem}
\usepackage{xcolor}
\usepackage{url}

\copyrightyear{2026}
\acmYear{2026}
\setcopyright{cc}
\setcctype{by}
\acmConference[SIGGRAPH Conference Papers '26]{Special Interest Group on Computer Graphics and Interactive Techniques Conference Conference Papers}{July 19--23, 2026}{Los Angeles, CA, USA}
\acmBooktitle{Special Interest Group on Computer Graphics and Interactive Techniques Conference Conference Papers (SIGGRAPH Conference Papers '26), July 19--23, 2026, Los Angeles, CA, USA}
\acmDOI{10.1145/3799902.3811135}
\acmISBN{979-8-4007-2554-8/2026/07}

\newcommand{\cmark}{\ding{51}}
\newcommand{\xmark}{\ding{55}}

\newcommand{\rev}[1]{\textcolor{black}{#1}}

\usepackage[ruled]{algorithm2e} 

\SetAlFnt{\small}
\SetAlCapFnt{\small}
\SetAlCapNameFnt{\small}
\SetAlCapHSkip{0pt}

\begin{document}

\title{R-DMesh: Video-Guided 3D Animation via Rectified Dynamic Mesh Flow}

\author{Zijie Wu}
\orcid{0009-0004-1171-9763}
\affiliation{%
  \institution{Huazhong University of Science and Technology}
  \city{Wuhan}
  \country{China}
}
\affiliation{%
  \institution{Tencent Hunyuan}
  \city{Shanghai}
  \country{China}
}
\email{zjw1031@hust.edu.cn}

\author{Lixin Xu}
\authornote{Project lead.}
\orcid{0009-0004-3579-3052}
\affiliation{%
  \institution{Tencent Hunyuan}
  \city{Shanghai}
  \country{China}
}
\email{21815063@zju.edu.cn}

\author{Puhua Jiang}
\orcid{0009-0001-1662-0736}
\affiliation{%
  \institution{Tencent Hunyuan}
  \city{Shenzhen}
  \country{China}
}
\email{puhuajiang@tencent.com}

\author{Sicong Liu}
\orcid{0009-0003-3652-1836}
\affiliation{%
  \institution{Tencent Hunyuan}
  \city{Shenzhen}
  \country{China}
}
\email{sicongliu@qq.com}

\author{Chunchao Guo}
\authornote{Corresponding author.}
\orcid{0009-0001-7465-802X}
\affiliation{%
  \institution{Tencent Hunyuan}
  \city{Shenzhen}
  \country{China}
}
\email{chunchaoguo@gmail.com}

\author{Xiang Bai}
\authornotemark[2]
\orcid{0000-0002-3449-5940}
\affiliation{%
  \institution{Huazhong University of Science and Technology}
  \city{Wuhan}
  \country{China}
}
\email{xbai@hust.edu.cn}

\renewcommand\shortauthors{Wu, Z. et al}

\begin{abstract}
Video-guided 3D animation holds immense potential for content creation, offering intuitive and precise control over dynamic assets. However, practical deployment faces a critical yet frequently overlooked hurdle: the pose misalignment dilemma. In real-world scenarios, the initial pose of a user-provided static mesh rarely aligns with the starting frame of a reference video. Naively forcing a mesh to follow a mismatched trajectory inevitably leads to severe geometric distortion or animation failure. To address this, we present Rectified Dynamic Mesh (R-DMesh), a unified framework designed to generate high-fidelity 4D meshes that are ``rectified'' to align with video context. Unlike standard motion transfer approaches, our method introduces a novel VAE that explicitly disentangles the input into a conditional base mesh, relative motion trajectories, and a crucial rectification jump offset. This offset is learned to automatically transform the arbitrary pose of the input mesh to match the video's initial state before animation begins. We process these components via a Triflow Attention mechanism, which leverages vertex-wise geometric features to modulate the three orthogonal flows, ensuring physical consistency and local rigidity during the rectification and animation process. For generation, we employ a Rectified Flow-based Diffusion Transformer conditioned on pre-trained video latents, effectively transferring rich spatio-temporal priors to the 3D domain. To support this task, we construct Video-RDMesh, a large-scale dataset of over 500k dynamic mesh sequences specifically curated to simulate pose misalignment. Extensive experiments demonstrate that R-DMesh not only solves the alignment problem but also enables robust downstream applications, including pose retargeting and holistic 4D generation. 
\rev{Code and pre-trained weights will be available at: \url{https://github.com/Tencent-Hunyuan/R-DMesh}.}
\end{abstract}




\begin{teaserfigure}
  \includegraphics[width=\textwidth]{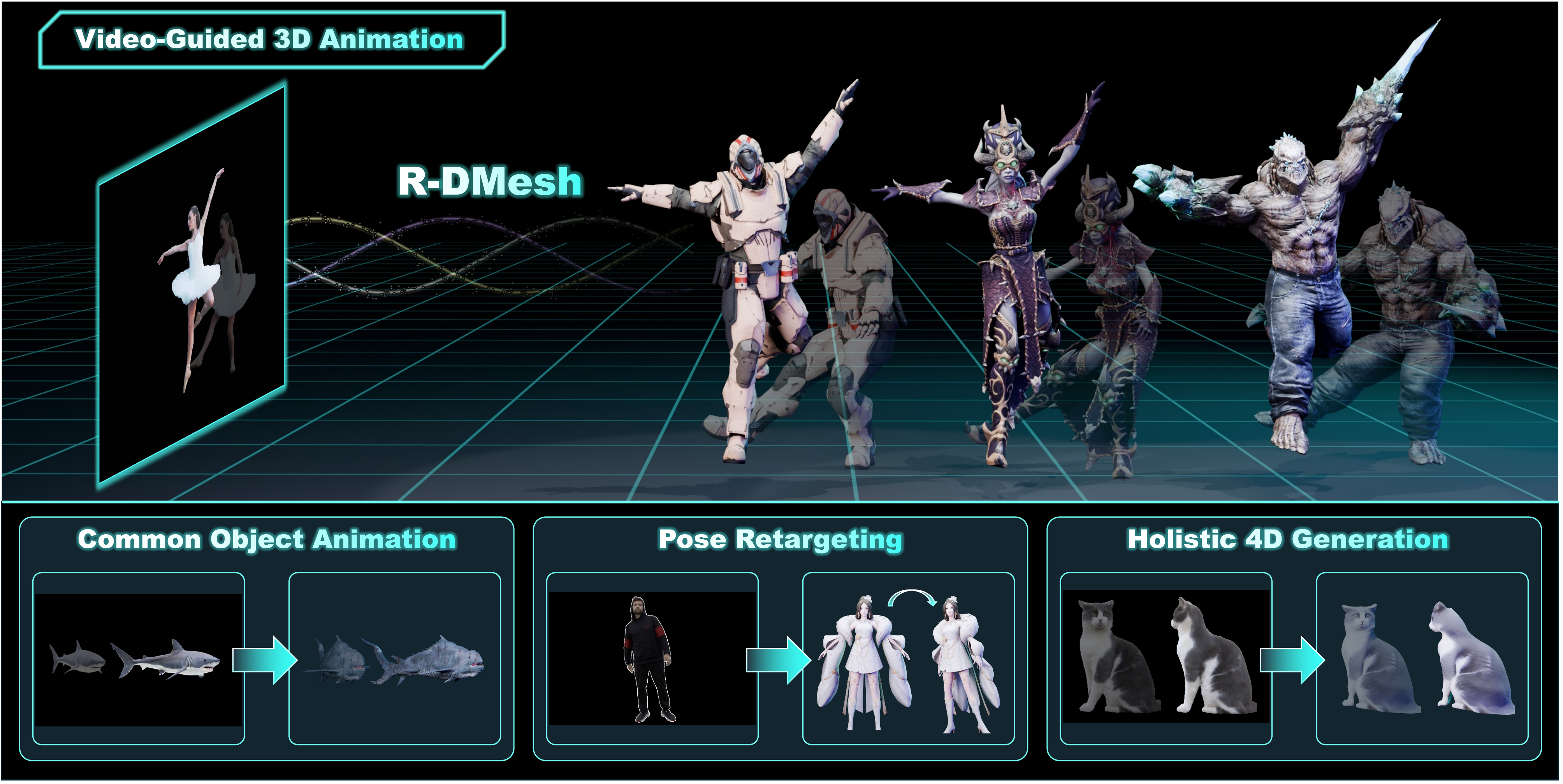}
  \caption{Video-Guided 3D Animation via Rectified Dynamic Mesh (R-DMesh). Given a monocular reference video (left), our method synthesizes high-fidelity, motion-aligned 4D meshes. Unlike traditional methods relying on skeletal rigging, R-DMesh directly predicts vertex trajectories, enabling the animation of topological varying characters (top) and common objects (bottom left) without explicit shape priors. Trained on same-identity video-4D pairs, our framework generalizes robustly to drive 3D models using reference videos from different identities, including in-the-wild footage. Furthermore, our approach supports versatile applications such as pose retargeting (bottom center) and holistic video-to-4D generation (bottom right).}
  \label{fig:teaser}
\end{teaserfigure}

\maketitle

\section{Introduction}
\label{sec:intro}
While generative models have revolutionized static 3D content creation~\cite{tripo,clay,hunyuan3d}, extending this success to the 4D domain remains formidable. The scarcity of high-quality 4D data, coupled with the immense complexity of modeling joint spatio-temporal distributions, hinders the development of generalizable 4D generative models.

Existing solutions typically fall into two categories. Holistic 4D generation approaches, which rely on SDS~\cite{sds} or multi-view video generation, struggle with spatio-temporal consistency. Without explicit 4D supervision, they often suffer from temporal flickering and fail to preserve local structural details, especially in unseen views. In parallel, Mesh-based Animation methods leverage high-quality static 3D assets. While promising, they are often confined to specific templates (e.g., SMPL~\cite{smpl}), require costly per-scene optimization, or lack precise control mechanisms, limiting their applicability to diverse, open-world objects.

\begin{figure}[t]
  \centering
   \includegraphics[width=1.0\linewidth]{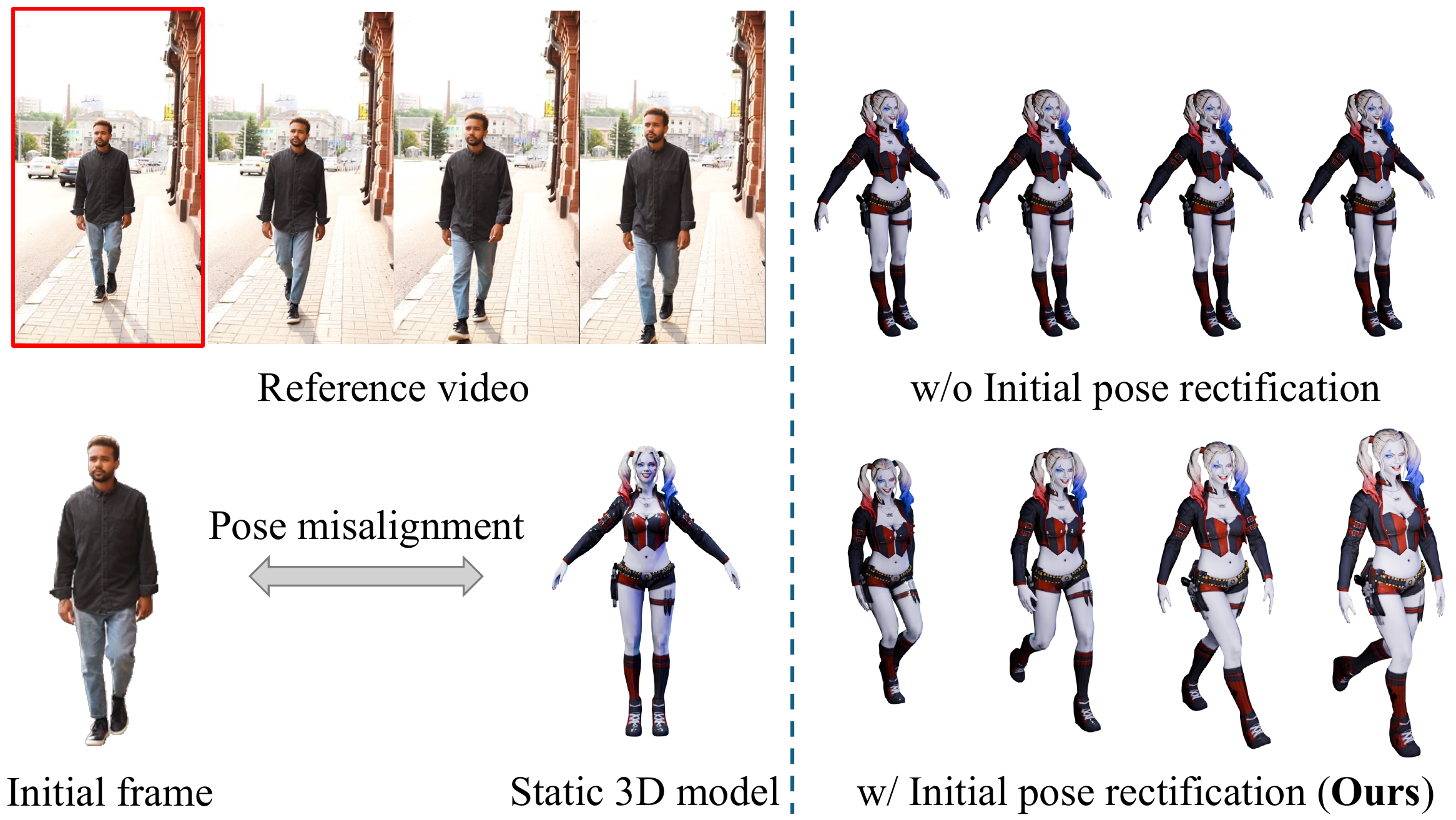}
   \caption{The challenge of pose misalignment in video-guided 3D animation. Significant discrepancies often exist between the initial video frame and the input 3D model. Directly transferring motion without addressing this leads to mismatched deformations or static outputs (top right). In contrast, our method performs pose rectification prior to motion transfer, laying a solid foundation for high-fidelity video-guided 3D animation (bottom right).}
   \label{fig:abl_motivation}
   \vspace{-.1in}
\end{figure}

To overcome these limitations, we advocate for Dynamic Mesh (DMesh) as the ideal representation, disentangling motion from geometry to leverage the high fidelity of 3D assets while focusing the learning capacity solely on motion dynamics. Furthermore, we identify monocular video as the most intuitive and information-rich signal for precise motion control.
However, a critical yet overlooked hurdle in video-guided 3D animation is the pose misalignment dilemma, as illustrated in Fig.~\ref{fig:abl_motivation}. In practical scenarios, the initial pose of a user-provided mesh rarely matches the starting pose of a reference video. Naively forcing a mesh to follow a mismatched video leads to severe distortion or animation failure. Therefore, a robust system must not only generate motion but also ``rectify'' the mesh to align with the video's context before animation begins.

To address these challenges, we present \textbf{Rectified Dynamic Mesh (R-DMesh)}, a unified framework for generating high-fidelity, motion-aligned 4D meshes. Our approach hinges on two core designs. First, we propose a novel R-DMesh VAE that disentangles the input into a conditional mesh, vertex jump offsets, and relative motion trajectories. The jump offsets are the key to our rectification capability, learning to transform the arbitrary pose of a conditional mesh to align precisely with the video's start frame. To coordinate these components, we introduce a Triflow Attention mechanism, which utilizes vertex-wise geometric features to modulate the three decoupled flows, enforcing physical priors of local rigidity and motion synergy. Second, for generation, we employ a Diffusion Transformer (DiT)~\cite{dit} architecture based on Rectified Flow~\cite{rf}. We condition the 4D DiT on video latent extracted from a pre-trained video generation model~\cite{wan}. By harnessing the rich spatio-temporal priors inherent in large-scale video models, we achieve coherent 4D generation with significantly reduced computational overhead.


To facilitate the above training process, we introduce Video-RDMesh, a large-scale dataset comprising over 500k high-fidelity dynamic sequences derived from Objaverse~\cite{objaverse, objaverse-xl}. Designed to simulate real-world inference scenarios (e.g., handling pose misalignment), this dataset provides a robust foundation for our framework. Consequently, our approach extends beyond standard video-guided 3D animation to support diverse downstream applications, including pose/motion retargeting and holistic 4D generation (as shown in Fig.~\ref{fig:teaser}), positioning it as a versatile solution for high-quality dynamic 3D content creation.

\begin{itemize}[leftmargin=1em]
\item We propose \textbf{R-DMesh}, a unified framework for video-guided 3D animation that effectively resolves the critical pose misalignment dilemma. By explicitly formulating a learnable jump offset, our method enables the seamless animation of static meshes with arbitrary initial poses using unaligned monocular videos.
\item We design a novel VAE architecture incorporating the Triflow Attention mechanism. By leveraging vertex-wise geometric features to modulate the disentangled flows of base geometry, jump offsets, and motion trajectories, this mechanism enforces physical priors of local rigidity and motion synergy, significantly enhancing generation fidelity.
\item We develop a scalable generative pipeline employing a Rectified Flow-based DiT conditioned on pre-trained VDM's latents. This design efficiently transfers rich spatio-temporal priors from video generation models to the 4D mesh domain, ensuring temporal coherence while reducing computational overhead.
\item We construct Video-RDMesh, a large-scale dataset comprising over 500k high-quality dynamic sequences with paired misalignment simulation. This dataset not only facilitates robust training but also empowers our framework to support diverse applications, including pose/motion retargeting and holistic 4D generation.
\end{itemize}
\section{Related Works}

\subsection{Holistic 4D Generation}
Due to the scarcity of 4D data, pioneering approaches~\cite{consistent4d, sc4d, 4dfy, mav3d} focus on distilling spatio-temporal priors from pre-trained image~\cite{zero123,zero123++}, video~\cite{zeroscope,svd}, or 3D generative models~\cite{lrm, lgm} to facilitate 4D generation. Typically, these methods employ Score Distillation Sampling (SDS)~\cite{sds} to compute gradients on renderings from specific viewpoints, which are then back-propagated to update the parameters of the underlying 4D representations (e.g., Dynamic NeRFs~\cite{dnerf} or 4DGS~\cite{4dgs}). Despite demonstrating the feasibility of 4D generation, the stochastic nature of SDS often leads to compromised spatio-temporal consistency. \rev{V2M4~\cite{V2M4} adopts a multi-stage scheme to optimize a dynamic mesh } However, the requirement for per-scene optimization significantly hinders the practical scalability of these methods.

To improve consistency, recent works~\cite{animate3d, diffusion4d, 4diffusion} propose synthesizing multi-view videos of dynamic objects.  However, these methods require an additional lifting step to reconstruct 4D representations. Moreover, without genuine 4D supervision, rendering from viewpoints outside the generated video's scope often leads to instability and artifacts. 

\subsection{3D Animation}
Instead of generating 4D content from scratch, another paradigm animates existing static 3D meshes. These approaches decouple motion from geometry, leveraging high-quality static 3D assets and requiring significantly less 4D training data. We categorize these methods into optimization-based and learning-based approaches.

Methods like~\cite{motiondreamer, animatetheuncaptured, ct4d} utilize SDS and reference-view losses to distill motion from video foundation models. While flexible, they suffer from the same bottlenecks: slow per-scene optimization and artifacts caused by the stochastic nature of SDS.
To achieve efficient inference, recent approaches focus on training feed-forward models.
While some methods~\cite{motiondiffuse, genhumanmotion, momask}~\rev{\cite{NeuralCapture}} have achieved impressive results on parametric models (e.g., SMPL~\cite{smpl}) or specific skeletons, their generalizability is severely limited. They are unable to animate models beyond the scope of their training templates or skeletons, let alone drive general-category objects.
Recent approaches targeting general skeletons~\cite{mocapanything, anytop} predict joint movements but struggle with non-rigid objects (e.g., fluids, clothing) and require pre-defined skeletons, preventing end-to-end animation of arbitrary meshes.

Closely related to our work is AnimateAnyMesh~\cite{animateanymesh, aam++}, which devise a text-to-trajectory rectified flow model. While innovative, text conditioning suffers from ambiguity and is limited by the expressiveness of textual descriptions. In contrast, we utilize video as a deterministic driving signal, ensuring precise motion control and enabling motion retargeting. 
\rev{Alongside our work, DriveAnyMesh~\cite{Driveanymesh} and the concurrent Mesh4D~\cite{Mesh4D} also adopt video guidance for dynamic mesh generation.}
However, utilizing video as a driving signal introduces a unique challenge absent in text-driven methods: the pose misalignment between the static mesh and the reference video's initial frame. \rev{While these existing video-driven approaches largely overlook this critical issue,} We address \rev{it} by designing a specialized VAE and Rectified Flow framework, effectively resolving the discrepancy and streamlining the animation pipeline.

\section{Methodology}

\noindent We propose R-DMesh, a framework synthesizing high-fidelity motion dynamics while rectifying the mesh's initial pose. It comprises two components: the \textbf{R-DMesh VAE} (Sec.\ref{rmesh-vae}), which compresses vertex trajectories into a disentangled latent space handling pose misalignment, and the \textbf{R-DMesh RF Model} (Sec.\ref{rdmesh-rf}), a Rectified Flow generator conditioned on video foundation model latents.

\subsection{R-DMesh VAE}
\label{rmesh-vae}

\begin{figure*}[t]
  \centering
   \includegraphics[width=1.0\linewidth]{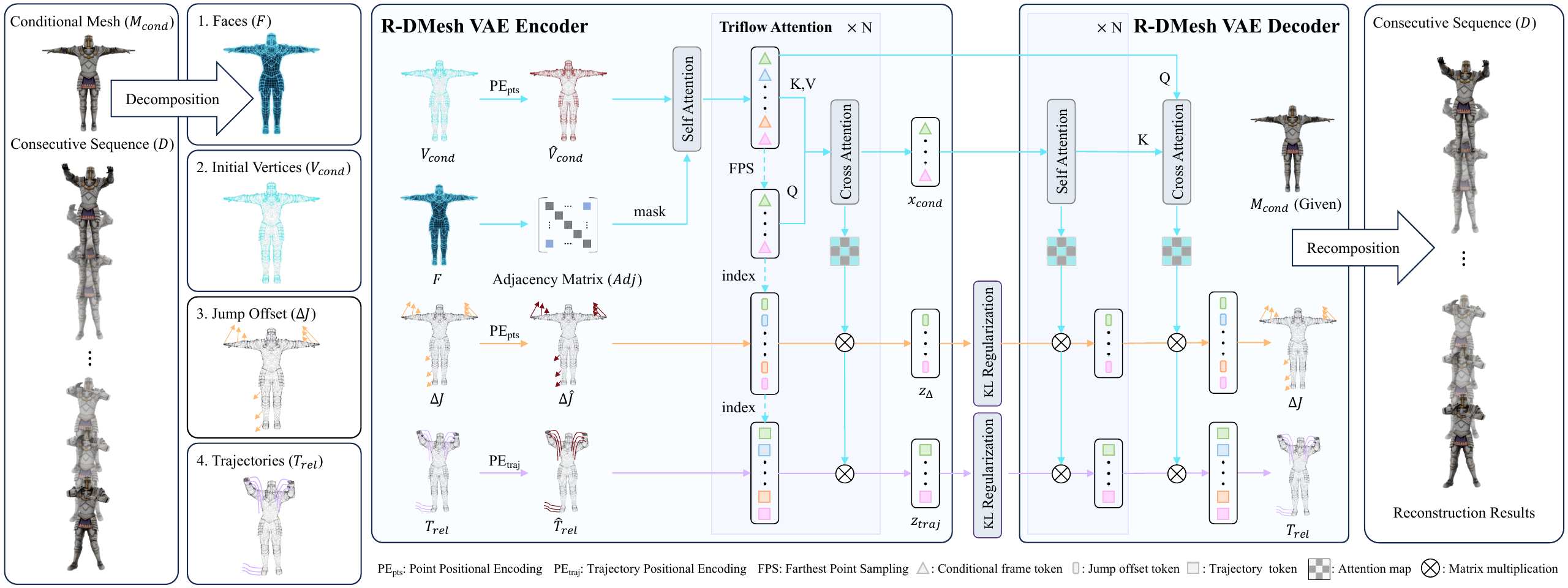}
   \caption{Illustration of our proposed \textbf{R-DMesh VAE}. It compresses and reconstructs dynamic mesh sequences conditioned on a static mesh of the same object in an arbitrary pose. (Left) Decomposition: The input sequence is decoupled into vertices $V_{cond}$, face $F$, global offsets $\mathit{\Delta } J$, and relative motion $T_{rel}$. 
(Middle) Encoder: The Triflow Attention mechanism jointly processes these components to capture spatio-temporal correlations, producing compact latent codes.
(Right) Decoder: The 4D sequence is reconstructed via a Triflow cross-attention, queried by the condition's vertex features.}
   \label{fig:rdmeshvae}
   \vspace{-.1in}
\end{figure*}

\noindent
To address the spatial discontinuity between a static conditional mesh $M_{cond}$ and a target dynamic sequence $D$, we introduce a hierarchical VAE that explicitly disentangles initial pose rectification from continuous dynamics (Fig.~\ref{fig:rdmeshvae}).


\paragraph{DMesh Decomposition and Representation.}
We consider a static conditional mesh $M_{cond} = (V_{cond}\in \mathbb{R}^{N \times 3}, F \in \mathbb{Z}^{M \times 3})$ and a target dynamic sequence $D= (V_{1:T}\in \mathbb{R}^{T\times N \times 3}, F \in \mathbb{Z}^{M \times 3})$, which share an identical topology with $N$ vertices and $M$ faces.

Direct reconstruction of the absolute vertex sequence $V_{1:T}$ presents significant challenges. First, modeling absolute coordinates inherently entangles the subject's intrinsic geometry with its motion dynamics. This forces the VAE to implicitly encode static geometric features within the motion distribution, unnecessarily increasing its complexity and hindering the convergence of subsequent generative models. Second, the spatial misalignment between the condition $V_{cond}$ and the sequence start state $V_1$ introduces abrupt discontinuities. Without explicit decomposition, these ``jumps'' contaminate the motion representation with static offsets, degrading temporal smoothness and complicating the learning of continuous dynamics.

To address these issues, we decompose the conditional mesh and target sequence into four disentangled components: the faces ($F\in\mathbb{Z}^{M\times 3}$), the initial vertices ($V_{cond}\in\mathbb{R}^{N\times 3}$), the jump offset ($\mathit{\Delta } J\in\mathbb{R}^{N\times 3}$), and the relative trajectories ($T_{rel}\in\mathbb{R}^{N\times (T\cdot3)}$), as illustrated in Fig.~\ref{fig:rdmeshvae}.
Specifically, the jump offset ($\mathit{\Delta } J$) and the relative trajectories ($T_{rel}$) are modeled as:
\begin{equation}
\mathit{\Delta } J = V_1 - V_{cond},~T_{rel} = V_{1:T} - V_1.
\end{equation}

Prior to this decomposition, we employ a canonicalization strategy. $V_{cond}$ is centered around its own centroid, whereas the target sequence $V_{1:T}$ is centered relative to the centroid of the first frame $V_1$. Subsequently, both components are normalized by the maximum absolute coordinate of the centered $V_{cond}$.
This strategy is necessitated by the lack of a corresponding control frame for $V_{cond}$ in the video-conditioning stage, which renders global translation unlearnable. By applying separate centering, we eliminate this non-essential displacement to stabilize inputs. Consequently, this effectively disentangles local alignment ($\mathit{\Delta } J$) from motion generation ($T_{rel}$).

\paragraph{Hierarchical Encoder with Triflow Attention.}
As illustrated in Fig.~\ref{fig:rdmeshvae}, we first project $V_{cond}$, $\mathit{\Delta } J$, and $T_{rel}$ into higher dimensions using positional encodings~\cite{3dshape2vecset} tailored for vertices and trajectories, respectively, yielding $\hat{V}_{cond}$, $\hat{\mathit{\Delta}} J$, and $\hat{T}_{rel}$. 
To encode local geometry, we perform self-attention on $\hat{V}_{cond}$ masked by the mesh adjacency matrix ($Adj$), inspired by ~\cite{animateanymesh}.

We employ Farthest Point Sampling (FPS)~\cite{pointnet++} on $\hat{V}_{cond}$ to determine $n$ representative indices, which are then used to gather corresponding feature subsets across all three modalities ($\hat{V}_{cond}^n, \mathit{\Delta} \hat{J}^n, \hat{T}_{rel}^n$).
Subsequently, we introduce \textbf{Triflow Attention} to aggregate global contexts. Instead of computing separate attention maps, we generate a shared attention map using the sampled geometry $\hat{V}_{cond}^n$ as the query and the full set $\hat{V}_{cond}$ as the key. This geometry-guided map is then applied to simultaneously aggregate features from all three modalities. By explicitly aligning motion and trajectory aggregation with geometric topology, this design encodes the intrinsic correlations between local rigidity and dynamics without compromising feature disentanglement, effectively reducing the learning difficulty:

\begin{equation}
\begin{aligned}
  A &= \text{Softmax}\left( \frac{\hat{V}_{cond}^n\cdot \hat{V}_{cond}^T}{\sqrt{d_k}  }\right), \\
  (\hat{V}_{cond}^n,\mathit{\Delta} \hat{J}^n,\hat{T}_{rel}^n) &= A\cdot (\hat{V}_{cond},\mathit{\Delta} \hat{J},\hat{T}_{rel}) + (\hat{V}_{cond}^n,\mathit{\Delta} \hat{J}^n,\hat{T}_{rel}^n). 
\end{aligned}
\end{equation}

We stack multiple Triflow Attention layers to strengthen the semantic information of each compressed token, ultimately obtaining the compressed vertex features $x_{cond}$, the jump features $x_{\mathit{\Delta}}$, and the relative trajectory features $x_{traj}$. Since our task targets mesh animation where $x_{cond}$ is always given, we only need to model the distributions of the probabilistic components. To this end, 
we model the probabilistic components $x_k \in \{x_\mathit{\Delta}, x_{traj}\}$ using Gaussian distributions $z_k = \mu_k + \sigma_k \cdot \epsilon_k$. The network is optimized using the standard KL divergence loss $\mathcal{L}_{kl_{k}}$ to regularize the latent space.



\paragraph{Decoder and Reconstruction.}
In the decoding phase, we project the three latent codes to a unified dimension and process them via stacked Triflow attention layers to reinforce spatial correlations. To propagate features back to the dense mesh topology, we employ a cross-attention mechanism using the fine-grained encoder features $\hat{V}_{cond}$ as queries and the processed latents $x_{cond}$ as keys. Finally, specific projection heads are used to reconstruct the jump offsets $\mathit{\Delta} J^{rec}$ and relative trajectories $T_{rel}^{rec}$, respectively.

To ensure stable convergence for both the jump offset and relative motion, we apply separate supervision to the reconstructed outputs. The reconstruction objectives are formulated as the Mean Squared Error (MSE) between the predicted and ground-truth values. The overall optimization objective for the R-DMesh VAE is a weighted sum of reconstruction and KL divergence losses:

\begin{equation}
\mathcal{L} = \mathcal{L}_{1}^{rec} + \mathcal{L}_{rel}^{rec} + \eta_1\cdot \mathcal{L}_{kl_\mathit{\Delta}} + \eta_{rel}\cdot \mathcal{L}_{kl_{traj}},
\end{equation}
where $\eta_1, \eta_{rel}$ are set to 1e-6 as default. Note that the reconstruction terms are averaged over all $N$ vertices.

\subsection{R-DMesh RF Model}
\label{rdmesh-rf} 

\begin{figure}[t]
  \centering
   \includegraphics[width=0.75\linewidth]{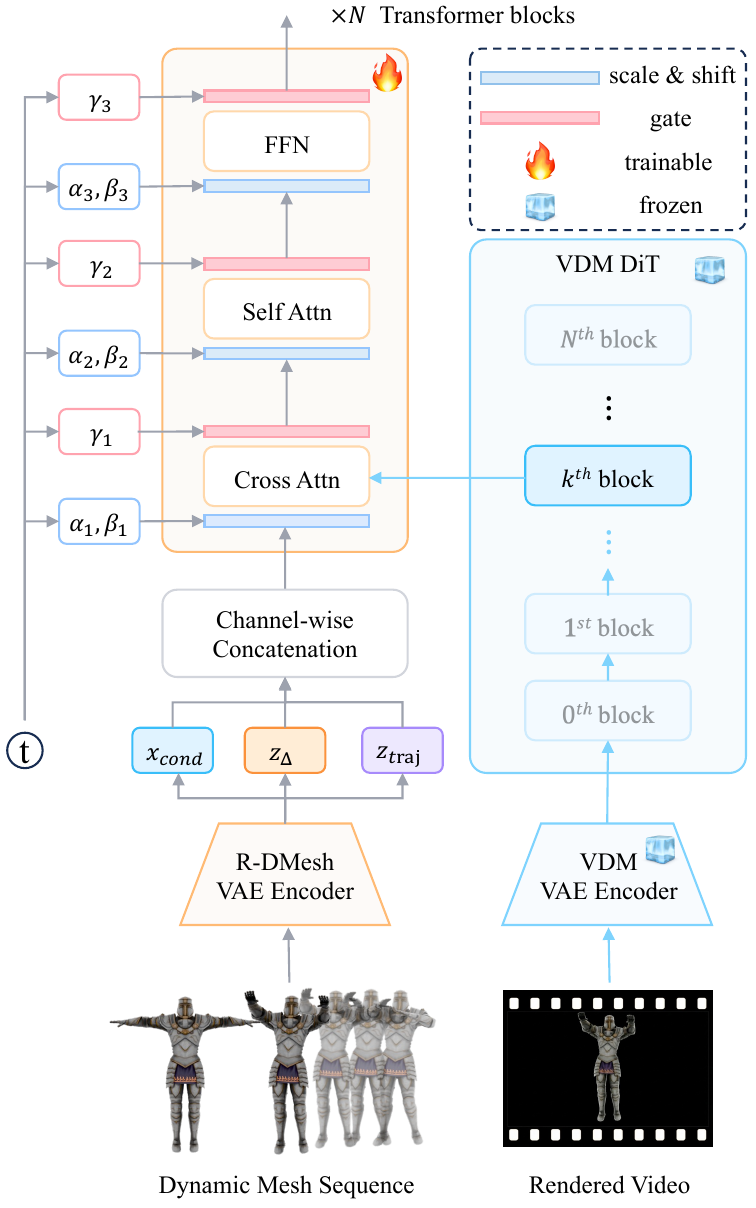}
   \caption{Illustration of our proposed \textbf{R-DMesh RF model}. We leverage a pre-trained, frozen Video Diffusion Model (VDM) as a strong visual prior. The VDM processes the reference video to extract rich semantic and dynamic features.
These features are injected into the trainable Transformer blocks via Cross-Attention, guiding the generation of mesh dynamics ($ z_{\Delta}, z_{traj}$).}
   \label{fig:rdmeshdit}
   \vspace{-.1in}
\end{figure}

\paragraph{Rectified Flow Formulation for Dynamic Meshes.}
We employ Rectified Flow~\cite{rf} to model the distribution of dynamic components $Z_{dyn} = [z_{\mathit{\Delta}}, z_{traj}]$ conditioned on the static structural latent $x_{cond}$ and video features $F_{vid}$. 
Since the initial mesh is given during inference, we treat the its representation $x_{cond}$ as a fixed clean condition.
During training, we construct the time-dependent state $Z_t$ via linear interpolation: $Z_t = t \cdot Z_{dyn} + (1-t) \cdot \epsilon$, where $\epsilon \sim \mathcal{N}(0, I)$ and $t \in [0, 1]$.
The network input is the channel-wise concatenation of the noisy dynamic state $Z_t$ and the clean condition $x_{cond}$, trained to minimize the flow matching objective:

\begin{equation}
  L_{RF} = \mathbb{E}_{t,Z_{dyn},\epsilon}[\left \| v_\theta([Z_t;x_{cond}],t,F_{vid}) - (Z_{dyn}-\epsilon) \right \|^2 ],
\end{equation}
where $F_{vid}$ represents the video guidance features.

\paragraph{Leveraging Pre-trained VDM Priors.}
To ensure the generated mesh motion aligns with the input video, we leverage a pre-trained Video Diffusion Model (VDM)~\cite{wan} as the feature extractor. We adopt the VDM for two main reasons:

First, generative models are trained on massive-scale video datasets to synthesize realistic pixels, forcing them to learn superior spatio-temporal correlations and physics priors. Second, modern VDMs employ a latent diffusion architecture that compresses video into compact spatio-temporal tokens. This significantly reduces computational costs while retaining richer semantic and dynamic information than pixel-space encoders.

Furthermore, we strategically extract features $F_{vid}$ from a specific intermediate layer, rather than the final output. We observe that shallower blocks primarily encode low-level spatial details, lacking global temporal integration. Conversely, the deepest layers, become overly specialized for the low-level denoising objective, thereby exhibiting diminished temporal consistency for motion dynamics. Consequently, we target intermediate blocks to capture rich spatio-temporal structures. 
These features are subsequently injected into our 4D branch via cross-attention.

\paragraph{Model Architecture.}
Our velocity estimator $v_\theta$ employs a DiT-based architecture specifically tailored for sequence modeling. As illustrated in Fig.~\ref{fig:rdmeshdit}, the network comprises $N$ stacked Transformer blocks, each containing a Cross-Attention layer, a Self-Attention layer, and a Feed-Forward Network (FFN).
To effectively inject the time step $t$ and stabilize training, we utilize AdaLN-Zero~\cite{cogvideox} modulation. The time embedding is projected to regress layer-specific scaling and shifting parameters ($\alpha, \beta$), as well as gating parameters ($\gamma$). Prior to each attention or FFN operation, input features are modulated via $\alpha$ and $\beta$, functioning as a time-dependent Adaptive Layer Normalization (AdaLN)~\cite{adaln}. Furthermore, the output of each sub-module is scaled by the learnable gate $\gamma$ before being added to the residual stream.
In terms of feature interaction, the input latents first aggregate motion cues from the video latents via cross-attention. Subsequently, self-attention layers model local motion dependencies, ensuring the generated motion is smooth and globally coherent. During training, we randomly drop the video conditioning latents with a probability of $p=0.1$ to enable Classifier-Free Guidance (CFG)~\cite{cfg} for flexible control during inference. This approach ultimately yields a high-fidelity, video-aligned network for video-guided mesh animation.

\section{Experiments}
\label{exp}

\subsection{Experimental Setup}

\paragraph{Implementation Details.}
Our R-DMesh VAE features a symmetric architecture with $8$ Triflow Attention layers in both the encoder and decoder. The encoder downsamples vertices by $8\times$ via Farthest Point Sampling (FPS) to generate queries. Input features (vertices, offsets, $64$-frame trajectories) are projected to $256$ dimensions and compressed into latents of sizes $64$, $16$, and $64$. The R-DMesh RF model comprises $12$ Transformer blocks with a dimension of $512$.
Training sequences are sliced into $64$-frame clips and filtered to exclude static samples (displacement $<0.01$).
Data samples are filtered to $<8,192$ vertices and a face-to-vertex ratio $<2.5$, then padded to fixed sizes ($8,192$ vertices, $20,480$ faces) during training. 
To improve robustness against pose discrepancies, we employ a misalignment simulation strategy by conditioning on a random frame from the sequence rather than the first frame. We condition the RF model using features from the $10$-th layer of the Wan2.2-TI2V-5B~\cite{wan} DiT, extracted from a $256 \times 256$ silhouette video rendering (tensor shape: $1088 \times 3072$). Training was performed on $32$ NVIDIA H20 GPUs: the VAE for $200$k iterations (cosine schedule, $2\text{e-}4 \to 2\text{e-}5$, $\sim54$h) and the RF for $300$k iterations (constant $1\text{e-}4$, $\sim120$h). Please refer to the $Suppl.$\footnote{$Suppl.$: the supplemental file} for further details.

\paragraph{Datasets and Evaluation Metrics.}
\label{dataset}
We train our model on Video-RDMesh, a large-scale dataset \rev{curated from \(252,823\) unique dynamic assets (largely rig-based animations) sourced from Objaverse~\cite{objaverse,objaverse-xl}. Through a pipeline of extraction, slicing, and rigorous filtering, we processed these assets to yield \(513,690\) high-quality, 64-frame vertex trajectory-based clips paired with reference videos.} \rev{To ensure visual consistency, the corresponding} ground-truth videos are rendered via Blender. \rev{Please refer to Sec.~1 of the supplemental file for detals of data curation.}
To benchmark performance in video-guided mesh animation, we introduce the Video-RDMesh test set. This dataset comprises two distinct subsets, each containing $64$ examples. The first subset consists of ground-truth (GT) dynamic mesh sequences paired with their corresponding frontal-view rendered videos. The second subset pairs conditional meshes with reference videos generated by WAN2.2-I2V-14B~\cite{wan}. For the latter, we synthesize reference videos by conditioning the video generator on front-view renderings of the input meshes and corresponding text prompts. The Video-RDMesh test set encompasses a diverse range of subjects, including humans, reptiles, flying creatures, and dynamic inanimate objects.

For evaluation, we employ specific protocols for each subset. For examples with GT dynamic meshes, we quantify performance using the average Euclidean Distance (EucD) of vertices relative to the GT at each timestamp. Conversely, for the subset utilizing generated reference videos, we render our results using identical camera settings and compute the frame-by-frame Peak Signal-to-Noise Ratio (PSNR) to assess visual alignment. Furthermore, we adopt the Subject Consistency (SC.) and Motion Smoothness (SM.) metrics from VBench~\cite{vbench} to evaluate the temporal consistency and motion fluidity of the generated 4D objects.

\begin{table}[t]
\centering
\caption{Quantitative comparison with state-of-the-art methods. We evaluate rendering quality (PSNR), temporal consistency (Subject Consistency "SC." and Motion Smoothness "SM." from VBench~\cite{vbench}), geometric accuracy (Euclidean Distance "EucD"), and inference time.}
\label{tab:comp}
\resizebox{0.8\linewidth}{!}{
\begin{tabular}{l|ccc|c|c}
\toprule
\multirow{2}{*}{Exp} & \multicolumn{3}{c|}{Rendering Metrics} & \multicolumn{1}{c|}{4D Metrics} & \multirow{2}{*}{Time ($\downarrow$)} \\ \cmidrule{2-5}
 & PSNR ($\uparrow$)  & SC. ($\uparrow$) & SM. ($\uparrow$) & EucD ($\downarrow$)  &  \\ \midrule
SC4D & 23.4   & 0.933 & 0.993 & / & $\sim$40m   \\
L4GM & 22.3   & 0.915 & \textbf{0.995} & / & $\sim$30s   \\
AAM & 13.5   & 0.948 & 0.994 & / &  \textbf{$\sim$8s}  \\
PUPT & 19.8   & 0.932 & 0.991 & 0.035 & $\sim$25m   \\ \midrule
Ours & \textbf{25.8}  & \textbf{0.949} & \textbf{0.995} & \textbf{0.012} &  $\sim$10s  \\ \bottomrule
\end{tabular}%
}
\vspace{-.1in}
\end{table}

\subsection{Comparisons}
\begin{figure*}[t]
  \centering
   \includegraphics[width=1.0\linewidth]{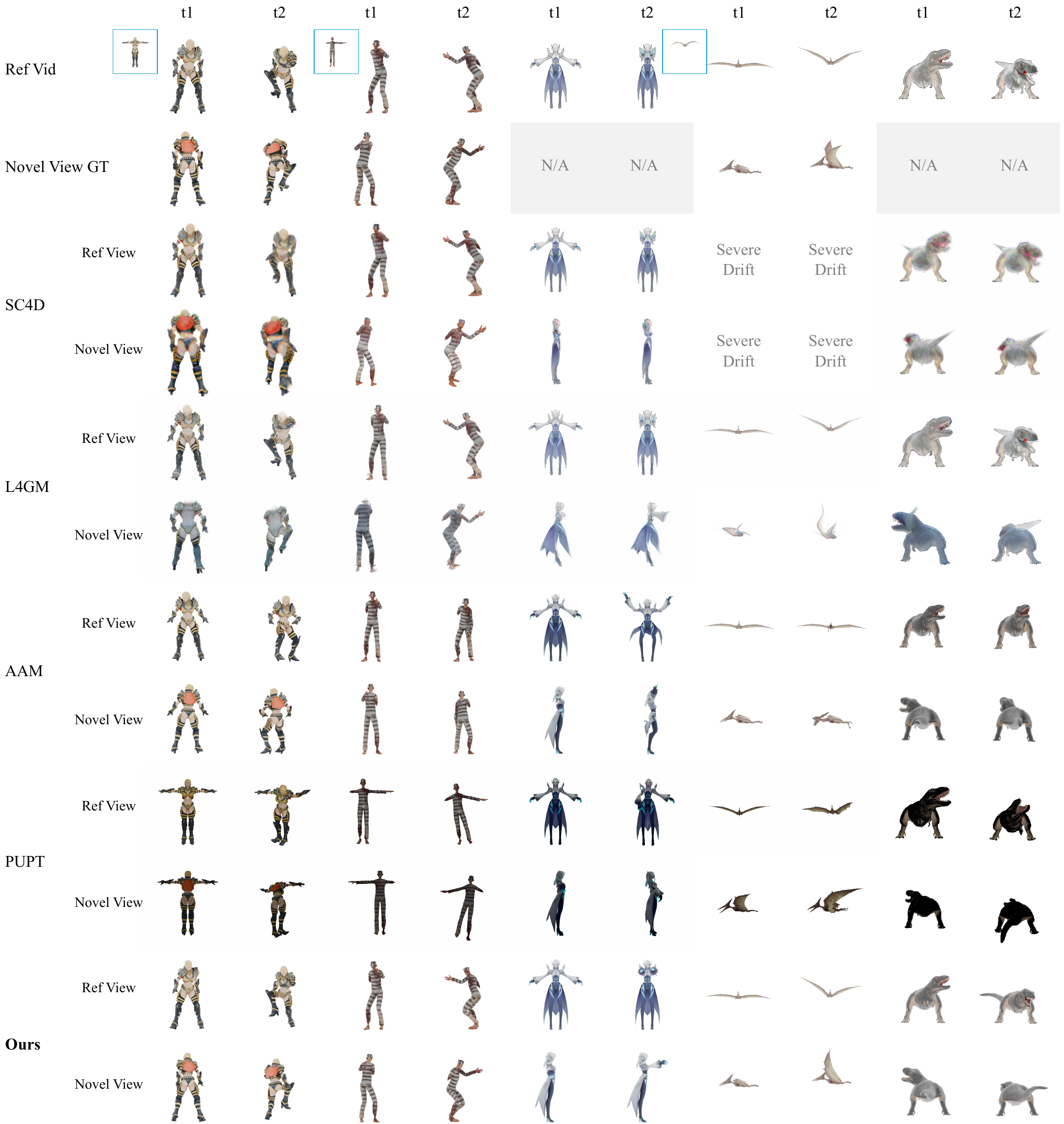}
   \caption{Qualitative comparison with state-of-the-art methods. We evaluate against video-to-4D methods (SC4D~\cite{sc4d}, L4GM~\cite{l4gm}), text-driven AAM (AnimateAnyMesh~\cite{animateanymesh}), and video-guided PUPT (Puppeteer~\cite{puppeteer}). Columns show renderings at initial ($t_1$) and random ($t_2$) timestamps across reference and novel views. Blue boxes highlight the input mesh's initial pose. "Severe Drift" marks an SC4D failure case due to excessive spatial deviation. "N/A" indicates missing ground truth for synthetic data. Zoom in for a better view.}
   \label{fig:comp}
\end{figure*}

Due to the scarcity of methods specifically targeting video-guided universal mesh animation, we compare our approach against four state-of-the-art methods from closely related tasks. These baselines include two video-to-4D generation methods, SC4D~\cite{sc4d} and L4GM~\cite{l4gm}, the text-driven mesh animation method AnimateAnyMesh~\cite{animateanymesh} (AAM), and the video-guided mesh animation method Puppeteer~\cite{puppeteer} (PUPT). While other methods directly utilize the reference video as a condition, for AAM, we employ Qwen-VL-2.5~\cite{qwen25} to caption the reference videos. These generated captions serve as the text prompts to ensure semantic alignment during the evaluation. We evaluate all methods using their official implementations on the datasets described in Sec.~\ref{dataset}, covering scenarios both with and without ground-truth 4D meshes. The qualitative and quantitative comparisons are presented in Fig.~\ref{fig:comp} and Tab.~\ref{tab:comp}, respectively.


\paragraph{Qualitative Comparison.}
As shown in Fig.~\ref{fig:comp}, while SC4D and L4GM generate renderings that closely align with the video in the reference view, they consistently exhibit color and shape distortions in novel views. This issue is particularly pronounced for reference videos featuring slender foreground objects (e.g., the pterosaur in Fig.~\ref{fig:comp}), where the generated results often suffer from severe deformation or drift significantly from the scene center, making optimization intractable. Although AAM generates high-quality animation results, it struggles to achieve fine-grained control via text prompts. Furthermore, for rare object categories, the learned distribution often fails to cover the corresponding motion manifold, leading to deformations or semantic motion artifacts (e.g., the pterosaur in Fig.~\ref{fig:comp}). PUPT, relying on 2D priors like optical flow, degrades in complex scenarios. It specifically fails in radial motions due to the lack of explicit depth (cols 5-6) and cannot handle pose misalignment between the mesh and the video's initial frame (cols 1-4).

In contrast, our proposed R-DMesh, trained on a large-scale dataset of high-quality video-DMesh pairs, generates dynamic mesh sequences that maintain high correspondence with the reference video while preserving global and local geometric fidelity. Benefiting from our sophisticated model design and curated training data, our method effectively resolves pose misalignment and ensures robust motion transfer. As demonstrated in Fig.~\ref{fig:comp}, our method significantly outperforms comparative approaches in terms of alignment with the reference video, shape preservation, and motion plausibility.

\paragraph{Quantitative Comparison.}
Tab.~\ref{tab:comp} summarizes the numerical results. We first evaluate the alignment fidelity between the rendered results and the reference videos. Our method achieves the highest PSNR ($25.8$), indicating that our driven meshes align most accurately with the target motion. regarding temporal consistency, our approach outperforms all baselines in both SC. ($0.949$) and SM. ($0.995$), ensuring that the generated 4D sequences maintain robust subject identity and fluid motion.

For geometric accuracy, we report the Euclidean Distance (EucD). Our method yields a significantly lower error ($0.012$) compared to PUPT ($0.035$), demonstrating more precise geometry deformation. In terms of efficiency, our approach takes only about $10$ seconds, which is comparable to the fastest baseline (AAM) and orders of magnitude faster than optimization-based methods. Overall, these quantitative metrics align well with the qualitative visualizations, further validating the superiority of our proposed framework.

\subsection{Ablation Studies}

\paragraph{R-DMesh VAE Components.}
We investigate four key designs in our VAE: (1) Dual-Norm, which normalizes the conditional mesh and the subsequent sequence independently; (2) Jump-Decomp, which explicitly models the displacement between the condition and the sequence start; (3) Tri-Attn, which disentangles the learning of geometry, pose, and motion; and (4) Decoup-Loss, which supervises offset and trajectory reconstruction separately.

\begin{figure}[t]
  \centering
   \includegraphics[width=1.0\linewidth]{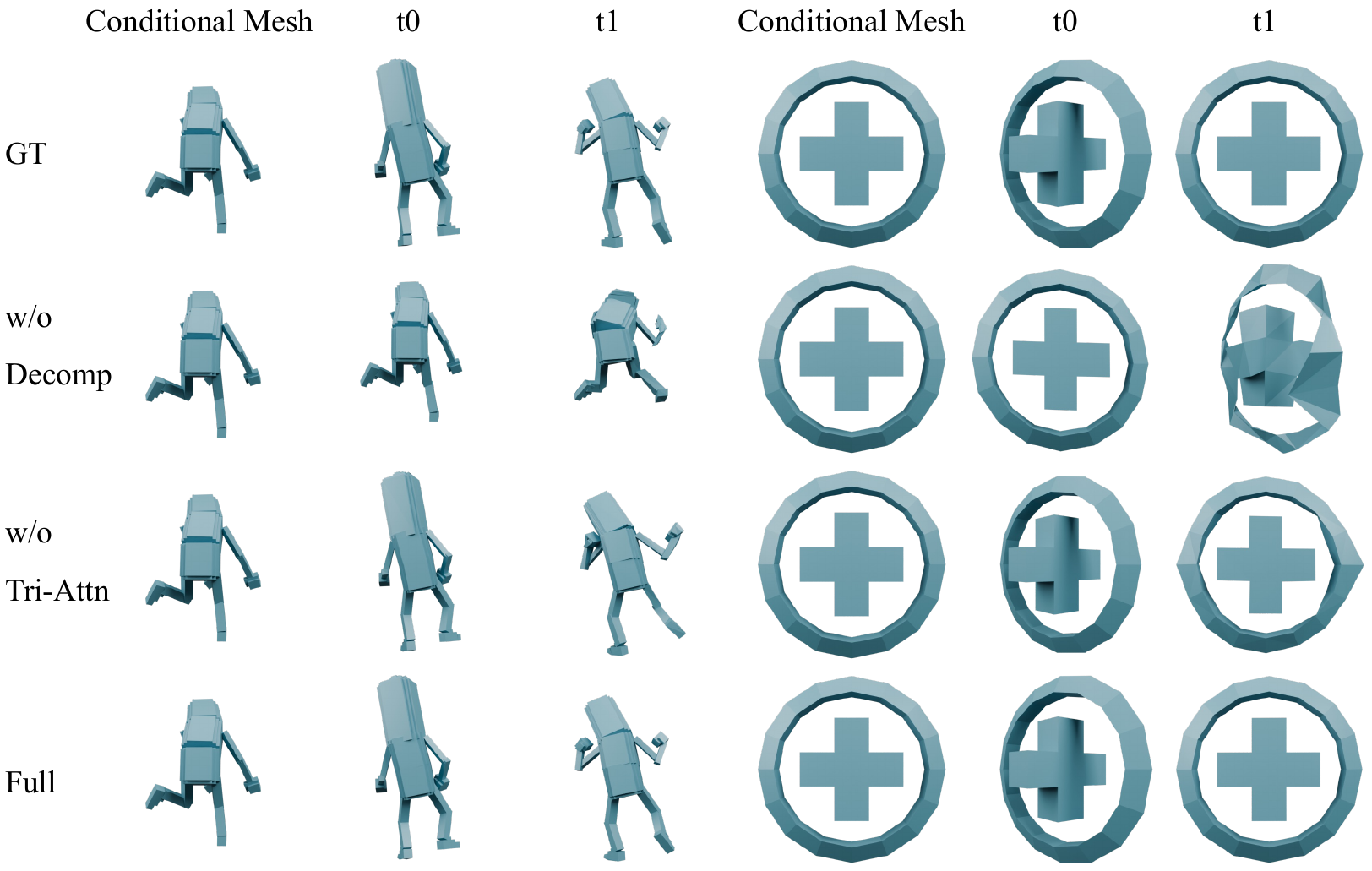}
   \caption{Visual ablation on Jump Decomposition and Triflow Attention. The start and end frame of the generated sequences are shown under $t_0, t_1$.}
   \label{fig:abl_vae}
   \vspace{-.1in}
\end{figure}

\begin{table}[t!]
  \centering
  \caption{Ablation studies of technical components. \textit{Dual-Norm}: Dual-Center Norm, \textit{Jump-Decomp}: Jump Offset Decomposition, \textit{Tri-Attn}: Triflow Attention, \textit{Decoup-Loss}: Decoupled Reconstruction Loss.}
  \resizebox{0.8\columnwidth}{!}{
    \begin{tabular}{cccc|c}
      \toprule
      \textit{Dual-Norm} & \textit{Jump-Decomp} & \textit{Tri-Attn} & \textit{Decoup-Loss} & EncD ($\downarrow$) \\
      \midrule
      \xmark & \cmark & \cmark & \cmark & 0.007 \\
      \cmark & \xmark & \cmark & \cmark & 0.025 \\
      \cmark & \cmark & \xmark & \cmark & 0.020 \\
      \cmark & \cmark & \cmark & \xmark & 0.007 \\
      \cmark & \cmark & \cmark & \cmark & \textbf{0.005} \\
      \bottomrule
    \end{tabular}
  }
  
  \label{tab:abl_vae}
  \vspace{-.1in}
\end{table}

Tab.~\ref{tab:abl_vae} quantifies the contribution of four core design elements in our R-DMesh VAE. The Jump-Decomp module emerges as the most pivotal factor. As evidenced in Tab.~\ref{tab:abl_vae} and Fig.~\ref{fig:abl_vae}, removing this module results in severe reconstruction degradation; notably, the generated initial frame ($t_0$) fails to capture the motion transition, remaining virtually identical to the conditional mesh pose. This failure occurs because, without explicit jump modeling, vertex trajectories are not de-centered. Consequently, the large coordinate offsets of the jump frame dominate the compression latent space, hindering effective trajectory clustering and separation. Furthermore, treating the jump frame indistinguishably during optimization neglects its foundational role; a poorly reconstructed initial state propagates errors, rendering the alignment of subsequent frames intractable.

The efficacy of Tri-Attn is also highlighted in our results. By leveraging priors of local rigidity and motion correlation, this module effectively disentangles the processing of jump vectors from relative motion trajectories. This isolation prevents mutual interference between these distinct information flows during optimization, significantly boosting reconstruction fidelity. Finally, Tab.~\ref{tab:abl_vae} confirms the positive impact of Dual-Norm and Decoup-Loss. Together, these components enable our full model to achieve high-fidelity compression and reconstruction for dynamic mesh sequences involving significant initial state transitions.

\paragraph{Video Feature Extraction.}
As outlined in Sec.~\ref{rdmesh-rf}, we leverage the pre-trained Wan2.2-TI2V-5B~\cite{wan} video model as our reference video feature extractor. To determine the optimal conditioning source, we conduct an ablation study on the feature layers. Given that Wan2.2-TI2V-5B comprises 30 transformer blocks, we experiment with latent outputs from the 0th (input), 10th, 20th, and 30th layers. These latents are injected into our R-DMesh RF model and trained for an identical number of iterations. As reported in Tab.~\ref{tab:abl_vid}, the latent output from the 10th layer proved most effective for our task. We also investigate concatenating latents from multiple layers; however, this approach yields inferior results compared to using the 10th layer alone. Consequently, we adopt the video features from the 10th transformer block of Wan2.2-TI2V-5B as the condition for our R-DMesh RF model.

\begin{table}[t]
  \centering
  \caption{Ablation study on video feature extraction layers. We utilize the Wan2.2-TI2V-5B model for feature extraction. DiT-L0, L10, L20, and L30 represent features from the corresponding intermediate blocks. }
  \label{tab:abl_vid}
  \resizebox{0.65\columnwidth}{!}{
    \begin{tabular}{cccc|c}
      \toprule
      DiT-L0 & DiT-L10 & DiT-L20 & DiT-L30 & EncD ($\downarrow$) \\
      \midrule
      \cmark &  &  &  & 0.059 \\
       & \cmark &  &  & \textbf{0.012} \\
       &  & \cmark & & 0.024 \\
       &  &  & \cmark & 0.037 \\
      \cmark & \cmark &  &  & 0.018 \\
       & \cmark & \cmark &  & 0.014 \\
       & \cmark &  & \cmark & 0.014 \\
      \bottomrule
    \end{tabular}
  }
  \vspace{-.1in}
  \label{tab:abl_vid}
\end{table}

\subsection{Application}
\label{sec:app}
Beyond the primary task of video-guided mesh animation, our framework exhibits significant versatility and potential for downstream tasks. We highlight three key applications: pose retargeting, motion retargeting, and holistic video-to-4D generation.

\begin{figure}[ht]
  \centering
   \includegraphics[width=1.0\linewidth]{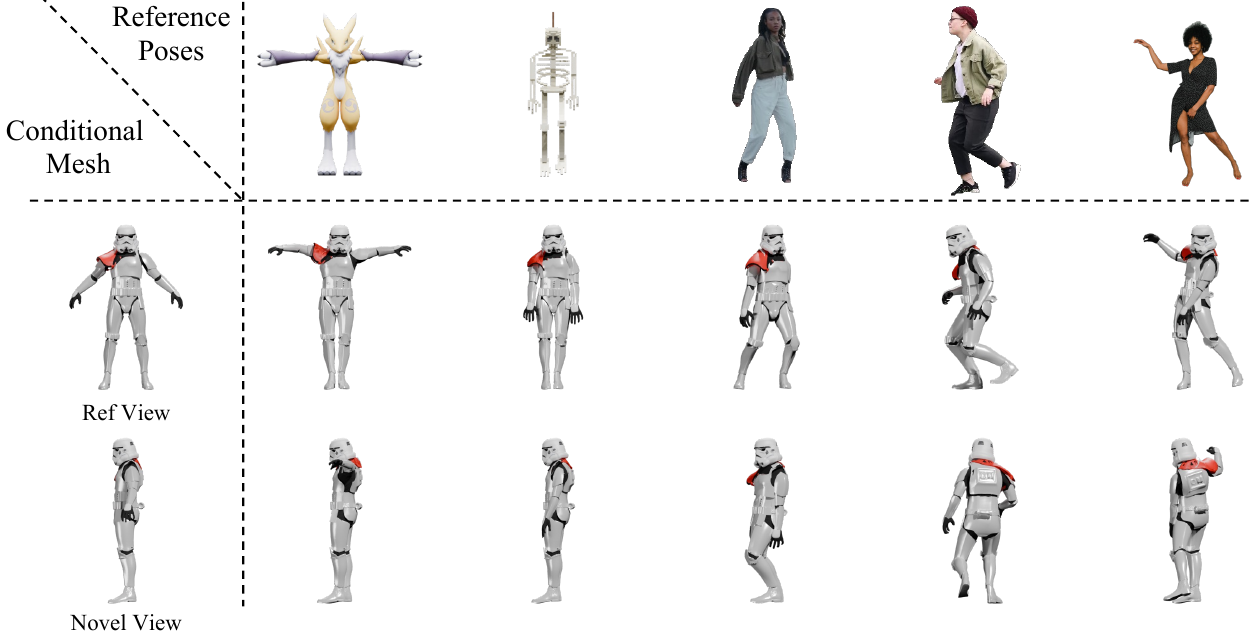}
   \caption{Pose retargeting application examples of our method.}
   \label{fig:app_repose}
\end{figure}

\paragraph{Pose Retargeting.} As demonstrated in Fig.~\ref{fig:app_repose}, our method robustly transfers poses from diverse sources to the conditional mesh. We showcase inputs ranging from synthetic 3D asset renderings (first two examples) to unconstrained in-the-wild images (last three examples). Even when presented with complex real-world poses, our approach successfully deforms the mesh to match the target posture. This robustness—particularly in handling the domain gap between synthetic and real data—is largely attributed to our explicit decoupling and modeling of vertex jump offsets, which effectively isolates large-scale pose changes from local details.

\begin{figure*}[t!]
  \centering
   \includegraphics[width=1.0\linewidth]{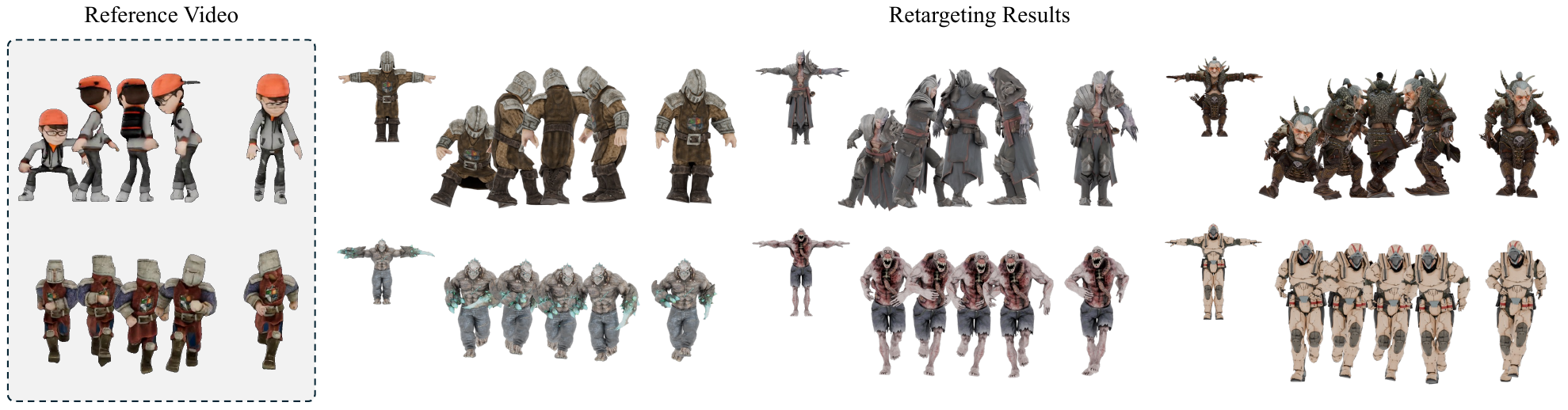}
   \caption{Motion retargeting application examples of our method. Left: Reference videos generated by video generation models. Right: Generated 3D animations. The conditional mesh (T-pose) is shown at the top-left of each sequence as the input condition, followed by the generated motion frames and the final pose.}
   \label{fig:app_retarget}
\end{figure*}

\begin{figure*}[ht]
  \centering
   \includegraphics[width=1.0\linewidth]{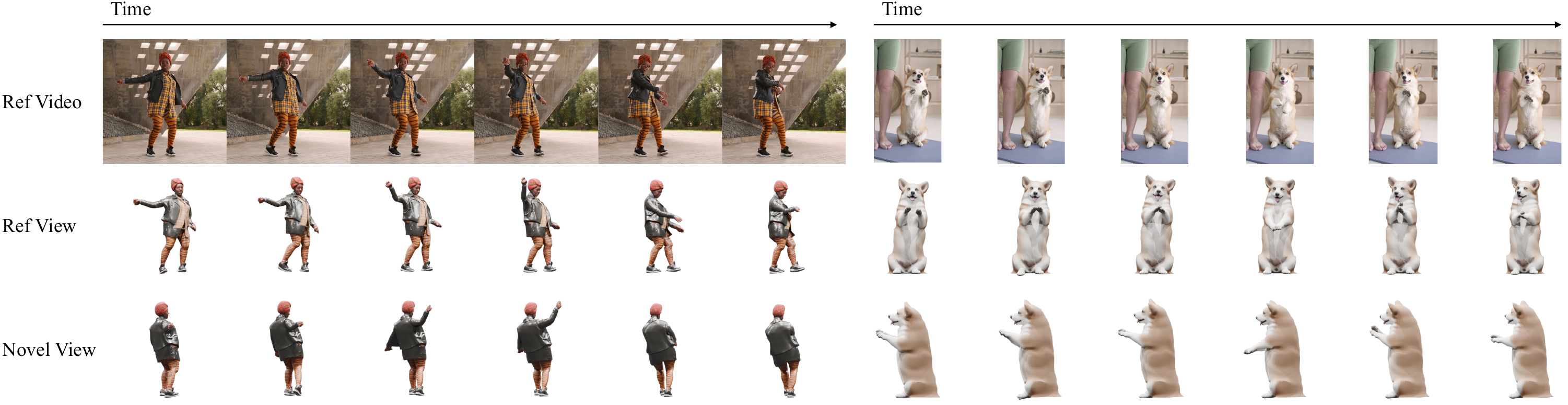}
   \caption{Holistic video-to-4D generation application examples of our method. The top row shows the reference videos. The middle row displays the reconstructed dynamic 3D content rendered from the reference camera view. The bottom row demonstrates the rendered novel views.}
   \label{fig:app_v24d}
\end{figure*}

\paragraph{Motion Retargeting.} Fig.~\ref{fig:app_retarget} illustrates our capability to transfer motion sequences from a driving video to a target 3D character. Our method generates highly consistent animation sequences that faithfully follow the driving motion while preserving the local geometry and identity of the target mesh. Notably, this is achieved without specific fine-tuning for different identities. The model generalizes well even when there are significant body shape discrepancies or complex accessories between the source and target. We attribute this zero-shot generalization capability to the high-quality distribution of our training data and the rich semantic priors inherited from the video generation model.

\paragraph{Holistic Video-to-4D Generation.} By integrating with state-of-the-art 3D generation models, our framework facilitates a complete video-to-4D pipeline. \rev{We utilize real-world videos as input, employing Hunyuan3D~\cite{hunyuan3d} to generate a static 3D mesh from the first frame as the conditional input. Using the segmented video~\cite{sam3} as guidance, our method then animates this static asset.} As shown in Fig.~\ref{fig:app_v24d}, this pipeline produces high-fidelity dynamic mesh sequences that are closely aligned with the reference video, successfully handling complex scenarios ranging from human dance moves to animal locomotion.

\begin{figure}[ht]
  \centering
   \includegraphics[width=1.0\linewidth]{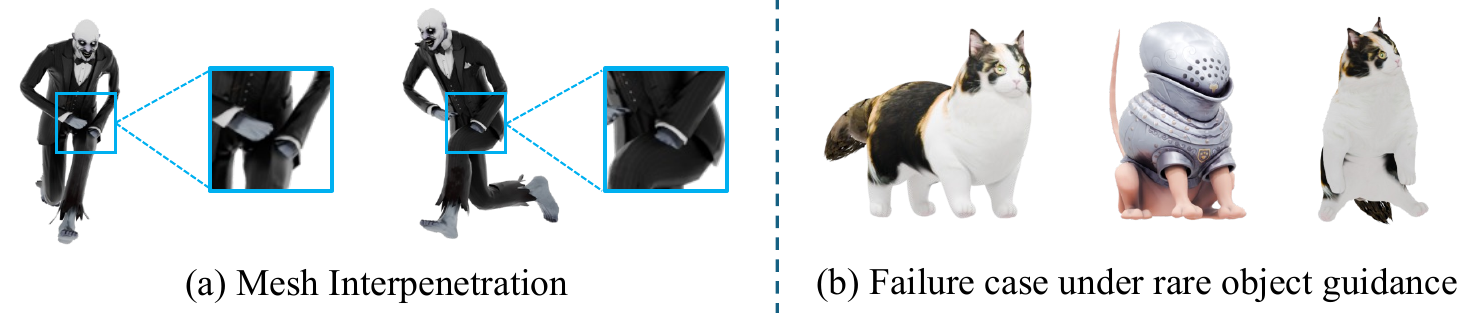}
   \caption{Limitations of our method. (a) Mesh Interpenetration. Our generated results may occasionally exhibit self-intersection artifacts. (b) Generalization to Rare Objects. When guided by rare objects (the armored cat), the plausibility of the generated pose and motion may be compromised.}
   \label{fig:limitation}
\end{figure}

\section{Limitation and Conclusion}
\paragraph{Limitation.}
\rev{As shown in Fig.~\ref{fig:limitation}}, our method currently faces two challenges: (1) Mesh Interpenetration: Occasional self-collisions appear in the output, attributed to noisy ground-truth data containing self-intersecting geometry that is difficult to fully purge. (2) Generalization to Rare Objects: The synthesis quality degrades under the guidance of rare categories, where data sparsity leads to unnatural deformations. Future work will focus on data filtering to eliminate intersections, post-hoc optimization, and dataset augmentation to improve robustness across diverse object categories.

\paragraph{Conclusion.}
In this work, we present \textbf{R-DMesh}, a robust framework for generating high-quality dynamic meshes controlled by monocular video. By identifying and addressing the pose misalignment dilemma—a pervasive issue in practical animation workflows, we introduce a disentangled representation that explicitly models the rectification process via a learnable jump offset. Our technical core, the Triflow Attention mechanism, successfully encodes physical priors into the generative process, ensuring that the synthesized motion remains geometrically consistent and locally rigid. Furthermore, by conditioning the Rectified Flow DiT on pre-trained video latents, we effectively bridge the gap between 2D video priors and 3D motion generation, achieving both high-fidelity and computational efficiency. The release of the Video-RDMesh dataset provides a substantial foundation for future research in this domain. We believe that R-DMesh not only advances the state of the art in video-guided 3D animation but also offers a versatile, general-purpose solution for creating diverse and coherent 4D assets.

\begin{acks}
This work is supported by the NSFC (62225603).
\end{acks}

\bibliographystyle{ACM-Reference-Format}
\bibliography{sample-bibliography}

\clearpage

\end{document}